# Private Meeting Summarization Without Performance Loss

Seolhwa Lee
Ubiquitous Knowledge Processing Lab (UKP Lab)
Department of Computer Science
Technical University of Darmstadt
Darmstadt, Hessen, Germany
whiteldark@korea.ac.kr

Anders Søgaard
Department of Computer Science
University of Copenhagen
Copenhagen, Denmark
soegaard@di.ku.dk## ABSTRACT
Meeting summarization has an enormous business potential, but in addition to being a hard problem, roll-out is challenged by privacy concerns. We explore the problem of meeting summarization under differential privacy constraints and find, to our surprise, that while differential privacy leads to slightly lower performance on in-sample data, differential privacy *improves* performance when evaluated on unseen meeting types. Since meeting summarization systems will encounter a great variety of meeting types in practical employment scenarios, this observation makes safe meeting summarization seem much more feasible. We perform extensive error analysis and identify potential risks in meeting summarization under differential privacy, including a faithfulness analysis.## CCS CONCEPTS
• **Computing methodologies** → **Natural language generation**; • **Security and privacy** → **Privacy protections**; • **Information systems** → **Summarization**.

**ACM Reference Format:**
Seolhwa Lee and Anders Søgaard. 2023. Private Meeting Summarization Without Performance Loss. In *Proceedings of the 46th International ACM SIGIR Conference on Research and Development in Information Retrieval (SIGIR '23), July 23–27, 2023, Taipei, Taiwan.* ACM, New York, NY, USA, 5 pages. https://doi.org/10.1145/3539618.3592042

## 1 INTRODUCTION
Professional meetings are often the pillars, if not the foundation, of the communication climate of businesses and organizations, and an integral component in management. Attendants typically deal with organizational matters, matters concerning the status of the profession, and scientific or technical developments. Meetings are a way for people to come up with ideas, share information, and make decision.

To capitalize on the ideas and decisions made in meetings, it is often important to produce minutes from such meetings. Minutes can be circulated to stakeholders that did not participate in the meeting, or be used to remind participants of the ideas and decisions made. Also, in the current work landscape, remote, online meetings have become much more common. This change in work culture comes with both challenges and opportunities. Since online meetings can be initiated in a more *ad hoc* manner, and may have a slightly more informal nature, it may be harder to produce standard minutes. On the other hand, such meetings are easier to record, and meeting transcripts can be produced through automatic speech recognition systems. For both physical and online meetings, meeting summarization is crucial for efficient dissemination of ideas and decisions, and for making this information available for later processing. Meeting summarization is already used in medical settings [8] and for product-related customer service [13].

The enormous business potential of meeting summarization is hampered by privacy concerns and data bottlenecks. Meeting transcripts can include private content and confidential information, including personally identifiable information, discussions of sensitive matters, or ideas and plans of considerable business value. Such information cannot be shared publicly, and models trained on such information may memorize and leak parts of it. Existing studies on text and meeting summarization mainly considered how to derive state-of-the-art result [9, 12], but no efforts for privacy concerns. To combat this issue, we explore the problem of meeting summarization under differential privacy (DP) [4, 5]. DP is a machine learning technique for the preservation of model privacy. With DP, models can be trained with a stochastic guarantee that the model will not leak training data.

The second challenge is the availability of representative training data. For privacy reasons, little public data is available, but more importantly, the meetings of organizations exhibit vary a lot in format and content. In practice, meeting summarization will almost always be a domain adaptation problem. Domain adaptation can be supervised or unsupervised, depending on whether the organization in question has available training data. We consider unsupervised domain adaptation below.

***Contributions.*** The combination of privacy requirements and domain differences present what seems like a formidable challenge. In this paper, we carry out a large set of cross-domain meeting summarization experiments under different privacy budgets. To our surprise, we find that training with differential privacy actually *improves* performance on unseen domains. We conduct a faithfulness evaluation and an extensive error analysis, showing that differentially private summaries are also more faithful. Our error analysis suggests that this is because the regularization effects of training with differential privacy prevents hallucination.

Permission to make digital or hard copies of all or part of this work for personal or classroom use is granted without fee provided that copies are not made or distributed for profit or commercial advantage and that copies bear this notice and the full citation on the first page. Copyrights for components of this work owned by others than the author(s) must be honored. Abstracting with credit is permitted. To copy otherwise, or republish, to post on servers or to redistribute to lists, requires prior specific permission and/or a fee. Request permissions from permissions@acm.org.
*SIGIR '23, July 23–27, 2023, Taipei, Taiwan*
© 2023 Copyright held by the owner/author(s). Publication rights licensed to ACM.
ACM ISBN 978-1-4503-9408-6/23/07...$15.00
https://doi.org/10.1145/3539618.3592042

Table 1: Performance comparison of differential privacy models (target $\epsilon = \{8\}$) and non-DP models. Each row (Pro., Com., Aca.) represents the training set and, each column represents the evaluation set. Also, B. denotes BERTScore. Note that performance denotes the best result so that it may be different $\epsilon$ on each model. Grey-colored background means improved/same performance than the baseline. Results are averaged over two random runs.

| Domain | Model & Method | | $\epsilon$ | Product | | | | Academic | | | | Committee | | | |
|---|---|---|---|---|---|---|---|---|---|---|---|---|---|---|---|
| | | | | R-1 | R-2 | R-L | B. | R-1 | R-2 | R-L | B. | R-1 | R-2 | R-L | B. |
| Pro. | GPT-2 | | - | 27.5 | 7.3 | 24.5 | 83.8 | 19.5 | 2.5 | 15.5 | 82.4 | 21.2 | 2 | 15.4 | 81.2 |
| | GPT2-m | | | 27.9 | 6.7 | 24.1 | 82.9 | 19.1 | 2.3 | 14.6 | 81.5 | 20.4 | 2.3 | 15.1 | 80.8 |
| | DialoGPT-m | | | 28.8 | 6.1 | 22.5 | 84 | 19.9 | 2.3 | 14.6 | 82.6 | 22.6 | 2.6 | 15.2 | 81.8 |
| | GPT-2 | DP-Ghost | 8 | 27.7 | 6 | 24.9 | 84.1 | 21.2 | 2.4 | 17.4 | 82.5 | 25.6 | 4.1 | 20.1 | 82.6 |
| | GPT2-m | | | 27 | 6 | 23.7 | 84.3 | 21.3 | 2.4 | 17.5 | 83 | 27.6 | 5 | 21.8 | 83.7 |
| | DialoGPT-m | | | 26.1 | 5.4 | 23.4 | 83.7 | 19.5 | 2.9 | 17.3 | 82.7 | 23.5 | 3.4 | 19.4 | 82.8 |
| | GPT-2 | DP-PFT | | 29.1 | 8.7 | 27.6 | 84.2 | 22.4 | 3.3 | 19.5 | 82.4 | 33.8 | 12.5 | 30.1 | 84.8 |
| | GPT2-m | | | 29.6 | 8.3 | 27.5 | 84.3 | 22.1 | 8.3 | 27.5 | 84.3 | 33.8 | 12.4 | 30.2 | 85.1 |
| Com. | GPT-2 | | - | 22.5 | 3.1 | 17.8 | 81.9 | 21.2 | 2.3 | 17.4 | 81.2 | 26.7 | 5 | 22.8 | 81.5 |
| | GPT2-m | | | 20.1 | 2.6 | 16.6 | 80.5 | 18.9 | 2 | 15.3 | 80.7 | 25.4 | 4 | 19.7 | 81.1 |
| | DialoGPT-m | | | 22.4 | 3.4 | 16.6 | 80.9 | 21.2 | 2.8 | 16.4 | 80.8 | 25.8 | 4.1 | 20.5 | 81.6 |
| | GPT-2 | DP-Ghost | 8 | 20.8 | 3.3 | 19.5 | 82.1 | 16.7 | 2.2 | 15.6 | 81.4 | 23.2 | 4 | 19.9 | 82.6 |
| | GPT2-m | | | 16.1 | 2.5 | 15.5 | 80.2 | 12.5 | 0.7 | 10.8 | 77.6 | 23.5 | 3.9 | 20.4 | 82.5 |
| | DialoGPT-m | | | 15.9 | 1.9 | 15.5 | 81.6 | 12.2 | 0.7 | 11.2 | 80.2 | 20.6 | 2.7 | 17.6 | 82.4 |
| | GPT-2 | DP-PFT | | 17.2 | 2.9 | 16.3 | 80.7 | 12.3 | 0.6 | 11.4 | 79.2 | 26.3 | 7.6 | 22.3 | 82.5 |
| | GPT2-m | | | 19.7 | 4.1 | 18.8 | 81.4 | 14.3 | 1.4 | 12.8 | 82 | 28.4 | 9.6 | 25.2 | 83.6 |
| Aca. | GPT-2 | | - | 22.7 | 3.5 | 18.5 | 80.7 | 20.6 | 3.6 | 17.6 | 81.9 | 22.9 | 2.9 | 18.1 | 80.1 |
| | GPT2-m | | | 21.8 | 3.6 | 20.1 | 79.8 | 21.3 | 4.2 | 21.5 | 79.7 | 23.5 | 3.2 | 20.6 | 79.2 |
| | DialoGPT-m | | | 23.4 | 3.6 | 18.5 | 80.8 | 20.8 | 3.8 | 18.3 | 81.5 | 22.5 | 3.2 | 18.3 | 79.8 |
| | GPT-2 | DP-Ghost | 8 | 21.5 | 3.6 | 20 | 82.9 | 20.5 | 2.4 | 18.2 | 83 | 24.9 | 3.4 | 20.2 | 83.1 |
| | GPT2-m | | | 22.1 | 3.9 | 20.4 | 83 | 18.4 | 2 | 16 | 81.6 | 24.3 | 4.1 | 19.8 | 82.8 |
| | DialoGPT-m | | | 22.4 | 3.5 | 18.9 | 82.1 | 18.9 | 2.2 | 16.8 | 81.1 | 20.7 | 2.7 | 16.4 | 81.9 |
| | GPT-2 | DP-PFT | | 19.6 | 3.4 | 17.6 | 81.6 | 16.6 | 2 | 15.3 | 80.7 | 26.8 | 8 | 23.1 | 83.1 |
| | GPT2-m | | | 20.9 | 4.6 | 19.2 | 82 | 16.7 | 1.8 | 15.1 | 81.3 | 25.6 | 7.2 | 21.9 | 82.9 |

## 2 METHOD

A meeting transcript $X = (x_1, x_2, ..., x_n)$ consists of $n$ turns of $x_i = (u_i, s_i)$, representing a speaker $s_i$'s utterance $u_i$. The aim of meeting summarization is generating a target summary $Y = (y_1, y_2, ..., y_m)$ by optimizing the conditional probability of $P(Y|X)$.

In this paper, we consider query-based meeting summarization in which every transcript-summary pair comes also with a query $Q = (w_1, ..., w_{|Q|})$, i.e., a sequence of words indicating what aspects of the transcript should be summarized. Our final goal of meeting summarization is thus reformulated to $P(Y|Q, X)$. We rely on the standard definition of $(\epsilon, \delta)$-differential privacy:

**Definition.** $(\epsilon, \delta)$-*differential privacy (DP)* [4]. *A randomized algorithm $Z$ is $(\epsilon, \delta)$-DP if for any two neighboring datasets $D$ and $D'$, which differ on a single element, and all subsets $S$ of possible outputs:*

$$Pr[Z(D) \in S] \leq e^\epsilon Pr[Z(D') \in S] + \delta.$$

If an algorithm satisfies DP, training it on similar training data will produce models that give similar predictions. The above definition generalizes differential privacy by introducing two privacy leakage parameters, $\epsilon$ and $\delta$. Intuitively, $\epsilon$ measures the degree of prediction differences we tolerate, and $\delta$ the number of exceptions we are willing to make (across the training set). *Small values imply stricter privacy guarantees.*

Our springboard for building DP private meeting summarization model is using pretrained language models trained with causal language modeling objectives on public data. We fine-tune pretrained language models with DP-Adam [1] in the above manner. DP has an expensive memory cost due to clipping per-example gradients [1, 2]. Li et al. [10] alleviate this issue by proposing a ghost clipping technique for saving memory. This makes it feasible to fine-tune larger language models under differential privacy. We also experiment with improving computational efficiency by not fine-tuning all parameters. Instead we follow Yu et al. [17] in fine-tuning only a small fraction of the total number of parameters. Training with DP-Ghost [10] enables us to train models with differential privacy in one hour on a single RTX 6000 GPU. Training with DP-PFT [17] enables us to train a bit slower, i.e., one and half hours on the same infrastructure.

## 3 EXPERIMENTS

### 3.1 Implementation Details

We use QMSum [20] to train and evaluate our query-based meeting summarization models. QMSum consists of meeting transcripts and summaries from three different domains. This enables us to perform cross-domain experiments. The data characteristics of QMSum are summarized in Table 2.

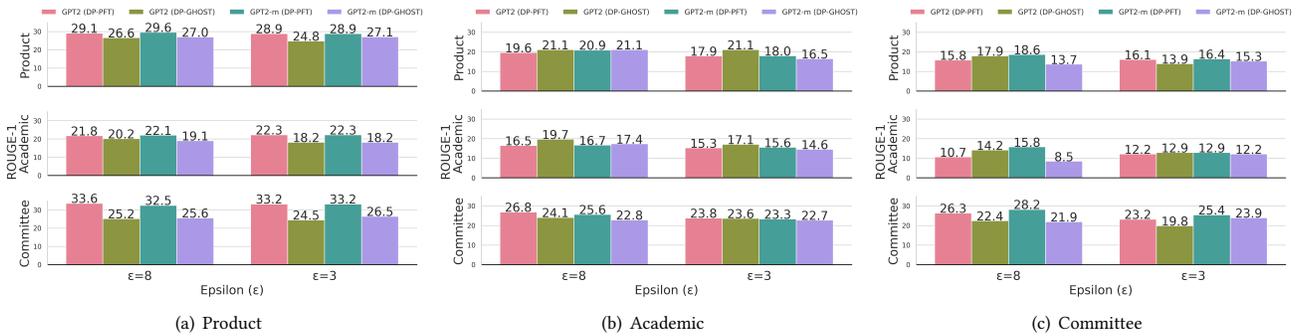

Figure 1: ROUGE-1 score of the last epoch depending on the epsilons ($\epsilon$)={3, 8}. Each (a), (b), and (c) denote the domain of the training set.

Table 2: Statistics of QMSum dataset according to domains. # pairs means query-summary pairs in train / valid / test set.

| Domain | Product (*pro*) | Academic (*aca*) | Committee (*com*) |
| --- | --- | --- | --- |
| # Meetings | 137 | 59 | 36 |
| # Pairs | 690 / 145 / 151 | 259 / 54 / 56 | 308 / 73 / 72 |
| Avg.len.meet | 6007.7 | 13317.3 | 13761.9 |
| Avg.len.sum | 70.5 | 53.7 | 80.5 |

For our baseline model, we fine-tune GPT2 [16] (including its smaller variant, GPT2-m(edium)) without differential privacy. As our second baseline, we fine-tune DialoGPT [19]. This model was pretrained on Reddit comments, i.e., conversational data that is more similar to meeting transcripts and its fit on this topic [6]. We now experiment with training differentially private models at different privacy levels $\epsilon \in \{3, 8\}$ and $\delta = 1/2|D_{train}|$, where $|D_{train}|$ indicates a training set size. These parameter settings were also used in Li et al. [10]. We use DP-Ghost [10] and DP-PFT [17]. We use ROUGE-1, 2, L [11] and BERTScore [18] as our performance metrics, comparing the similarity of the gold summaries and the predicted summaries. For our faithfulness analysis, we use ROUGE-L [11], comparing predictions to the input transcripts. All scores are F1 scores (see Appendix A for further details).

### 3.2 Results

In Table 1, we present the performance of our non-DP baseline models and our DP models across different cross-domain settings. DP typically struggles to achieve performance comparable to non-DP baselines [3], but our results, surprisingly, reflect an opposite trend. Training with differential privacy leads to *better* performance, especially in cross-domain settings. We grey-color and bold-face performance numbers in settings where training with differential privacy were superior to baseline public training.

***In-domain & out-of-domain***. In-domain performance of our DP models is competitive, sometimes better than our baseline models, especially in the Product domain. Differentially private models are better by some margin in the cross-domain setting, however. Training GPT2 models on Products and evaluating them in the Committee domain, we see absolute ROUGE-1 improvements of 4.4 points when training with DP-Ghost, and absolute ROUGE-L improvements of 4.7 points, for example. Training with DP-PFT is even better, leading to 12.6 and 14.7 points improvements, respectively. Generally, DP-PFT leads to somewhat larger improvements than DP-Ghost.

A few smaller observations are in order: DP models show poor performance when training on Committee data and evaluating on data from the Academic domain, but good performance in the opposite direction. We investigate this issue on error analysis. DialoGPT model shows slightly better performance than the GPT2-m model in the baseline setting, but differentially private training does not lead to similar gains for DialoGPT.

***Privacy levels***. We compare DP model performance across different privacy levels, varying $\epsilon$. Results are presented in Figure 1. We observe that looser privacy guarantees show better performance than strict guarantees, suggesting that the regularization effects of DP only lead to better performance at moderate privacy levels.

***Model size***. We see little to no differences in performance between GPT2 and GPT2-m, and they both seem to benefit equally from training with differential privacy.

## 4 ERROR ANALYSIS

In examining the predicted summaries, we observe differences in summary length for DP and non-DP models. See the plots in Figure 2. DP models exhibit a flatter, smoother distribution of summary lengths than non-DP models. In fact, the non-DP models seem to often predict summaries of the same length. This could indicate high degrees of memorization and hallucination.

For this reason, we investigate the faithfulness of DP and non-DP models by comparing the ROUGE-L scores between the input meeting transcripts and the generated output, i.e., the predicted summaries, in the test set to see the longest sequence overlaps for predictions of the two classes of models. See Table 3. We report numbers on validation data to minimize leakage. Consistently, the predicted summaries of our differentially private models are more faithful (bold numbers) than the baseline summaries. Also, aforementioned case of performance issues between Academic and Committee domains, we can confirm that training with Academic domain and evaluating on Committee domain has more word overlap than the opposite. To dig deeper and better understand this result, we manually analyze the predicted summaries. See Table 4 for an example. We find that DP tends to prevent hallucinations that occur in our baseline models. Our baseline summarization models often generate unwarranted phrases that are not contained in the original transcripts, e.g., *fruit of vegetable shape*. See also Appendix C.

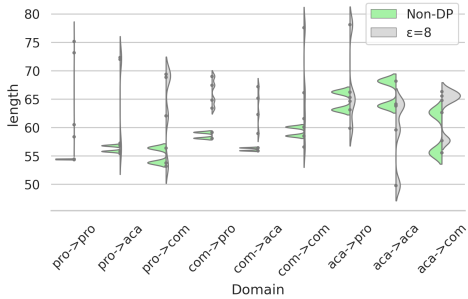

Figure 2: Average summary length for non-DP and DP ($\epsilon = 8$) GPT2 and GPT2-m models. The length of DP indicates the average length of both DP-Ghost and DP-PFT.

Table 3: ROUGE-L between input and prediction for Non-DP and DP GPT2 models.

| valid train | Non-DP | | | DP-Ghost ($\epsilon = 8$) | | |
|---|---|---|---|---|---|---|
| | P. | A. | C. | P. | A. | C. |
| P. | 16.2 | 9.5 | 10 | 17.1 | 11.7 | 11.8 |
| A. | 11.2 | 9.8 | 7.7 | 16.6 | 13.4 | 13.1 |
| C. | 10.2 | 8.6 | 10.6 | 16.5 | 12.2 | 14 |

Table 4: Example summaries.

| | Example (generated summary) |
|---|---|
| Non-DP | the market the remote control could make several different fronts, like the surf-board. the marketing put forward-looking remote control in a **fruit or vegetable shape**... |
| DP ($\epsilon = 8$) | industrial design project manager agreed that it would be easy to make a button which would be able to be attached to the remote control... |

## 5 CONCLUSION

We explore meeting summarization under differential privacy. We found that DP surprisingly leads to improved cross-domain performance, suggesting the two main challenges of meeting summarization–summarization and cross-domain robustness–can be solved in one go.

## ACKNOWLEDGEMENTS

This work was funded by the Innovation Fund Denmark through Grand Solutions grants PIN and AutoAI4CS.

## A HYPERPARAMETERS

For the DP-PFT model, we used LoRA (Low-Rank Adaptation) [7] for private fine-tuning. To determine the noise multiplier, we use Rényi differential privacy (RDP) [15]. Also, we used beam search for both models in the generation step. See Table 5 for the full list of hyperparameters[1].

Table 5: Hyperparameters for DP-Ghost and DP-PFT.

| Method | DP-Ghost | DP-PFT |
|---|---|---|
| DP guarantee ($\epsilon, \delta$) | ({3,8}, $1/{2|D_{train}|}$) | ({3,8}, $1/{2|D_{train}|}$) |
| Clipping norm | 0.1 | 0.0 |
| Batch size | 4 | 4 |
| Learning rate | 2e-3 | 4e-4 |
| Weight decay | 0.0 | 0.01 |
| Epoch | 20 | 20 |
| Optimizer | DP-Adam [1] | AdamW [14] |
| Beam size | 5 | 5 |
| Accounting mode | RDP | RDP |

## B LENGTH OF GENERATED SUMMARIES FROM DIALOGPT

Using the test set, we evaluated the generated summaries between non-DP and DP of DialoGPT-m. Unlike GPT2-related models, DP of DialoGPT-m produced shorter summaries than the non-DP model.

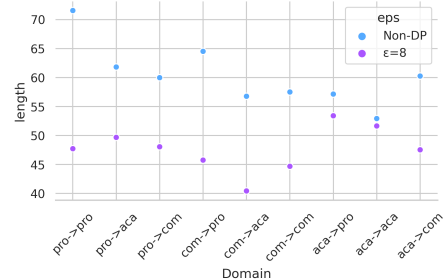

Figure 3: The average length of summaries with different epsilons (eps) from DialoGPT-m of Table 1's DP model.

## C GENERATED EXAMPLES

In the example below, † and ∗ show how non-DP models hallucinate outputs; bold words are unseen words in the source text. Generally, non-DP has more hallucination (blue) and less memorization (orange) than DP.

Table 6: Example summaries of continued from Table 4.

| | Example |
|---|---|
| Non-DP† | that they had been used in the work in the same time, and they would be a lot of the system. however, it was not put up to help them in the **whole-led approach** to the same step forward and have been working in the **baccol shillabeer** answered that they would have a number of wales. |
| DP ($\epsilon =8$)† | to drop the grade, they would not be able to drop it. I spoke earlier about the experience of my daughter's student who was permitted to drop one grade at the end of the course, and she was told that it was mixed. she told me that it depended on the students and the area that they were going to the cardiff metro university. the students were told that they could drop one or two grade, and they were told to drop two or three grades. |
| Non-DP∗ | the project manager proposed to make the remote control a **round shape instead of the traditional rectangular**, but user interface opposed this idea for fear that it would not be too small or be too big or too small, it would be fit into the hand. industrial checked this might increase the cost of the controller. |
| DP ($\epsilon =8$)∗ | it would be too hard to hold the remote in your hands when it was big and it would have to be big enough to hold it on your wrist. interface manager agreed that the remote would be smaller than the one in the remote, but it was still big enough for the user interface. |
| Meeting transcript‡ | ... project manager: yep. industrial designer: there are some uh hybrid hybrid devices which incorporate a **solar panel** and rechargeable batteries... |
| Non-DP‡ | in addition, the industrial designer had some responsive sparkles but other elements such as the cradle in a dark room. the project manager proposed to add a **solar cell** in the remote, which could control the cost of adding a child lock which could also be used as a cradle for the remote. this might not be a problem. |
| DP ($\epsilon =8$)‡ | marketing and marketing agreed that it would be a case for the cost of the **solar panel**. industrial user interface suggested that it could be used for the rechargeable battery. so it would cost less than the cost. industrial project manager agreed that the cost would be too expensive. and the team decided that it was too cheap. |

---

[1]https://github.com/coastalcph/private_meeting_summarization

## REFERENCES
[1] Martin Abadi, Andy Chu, Ian Goodfellow, H Brendan McMahan, Ilya Mironov, Kunal Talwar, and Li Zhang. 2016. Deep learning with differential privacy. In *Proceedings of the 2016 ACM SIGSAC conference on computer and communications security*. Proceedings of the 2016 ACM SIGSAC Conference on Computer and Communications Security, 308–318.
[2] Raef Bassily, Adam Smith, and Abhradeep Thakurta. 2014. Private empirical risk minimization: Efficient algorithms and tight error bounds. IEEE, 2014 IEEE 55th Annual Symposium on Foundations of Computer Science, 464–473.
[3] Christophe Dupuy, Radhika Arava, Rahul Gupta, and Anna Rumshisky. 2022. An Efficient DP-SGD Mechanism for Large Scale NLU Models. IEEE, ICASSP 2022-2022 IEEE International Conference on Acoustics, Speech and Signal Processing (ICASSP), 4118–4122.
[4] Cynthia Dwork, Frank McSherry, Kobbi Nissim, and Adam Smith. 2006. Calibrating noise to sensitivity in private data analysis. In *Theory of cryptography conference*. Springer, 265–284.
[5] Cynthia Dwork, Aaron Roth, et al. 2014. The algorithmic foundations of differential privacy. *Found. Trends Theor. Comput. Sci.* 9, 3-4 (2014), 211–407.
[6] Xiachong Feng, Xiaocheng Feng, Libo Qin, Bing Qin, and Ting Liu. 2021. Language Model as an Annotator: Exploring DialoGPT for Dialogue Summarization. In *Proceedings of the 59th Annual Meeting of the Association for Computational Linguistics and the 11th International Joint Conference on Natural Language Processing (Volume 1: Long Papers)*. Association for Computational Linguistics, Online, 1479–1491. https://doi.org/10.18653/v1/2021.acl-long.117
[7] Edward J Hu, Yelong Shen, Phillip Wallis, Zeyuan Allen-Zhu, Yuanzhi Li, Shean Wang, Lu Wang, and Weizhu Chen. 2021. Lora: Low-rank adaptation of large language models. *arXiv preprint arXiv:2106.09685* (2021).
[8] Anirudh Joshi, Namit Kataria, Xavier Amatriain, and Anitha Kannan. 2020. Dr. Summarize: Global Summarization of Medical Dialogue by Exploiting Local Structures.. In *Findings of the Association for Computational Linguistics: EMNLP 2020*. Association for Computational Linguistics, Online, 3755–3763. https://doi.org/10.18653/v1/2020.findings-emnlp.335
[9] Seolhwa Lee, Kisu Yang, Chanjun Park, Joao Sedoc, and Heuiseok Lim. 2021. Towards Syntax-Aware Dialogue Summarization using Multi-task Learning. Workshop on Widening NLP at EMNLP 2021.
[10] Xuechen Li, Florian Tramer, Percy Liang, and Tatsunori Hashimoto. 2021. Large language models can be strong differentially private learners. *arXiv preprint arXiv:2110.05679* (2021).
[11] Chin-Yew Lin. 2004. ROUGE: A Package for Automatic Evaluation of Summaries. In *Text Summarization Branches Out*. Association for Computational Linguistics, Barcelona, Spain, 74–81. https://aclanthology.org/W04-1013
[12] Hui Lin and Vincent Ng. 2019. Abstractive summarization: A survey of the state of the art. In *Proceedings of the AAAI conference on artificial intelligence*, Vol. 33. 9815–9822.
[13] Chunyi Liu, Peng Wang, Jiang Xu, Zang Li, and Jieping Ye. 2019. Automatic dialogue summary generation for customer service. In *Proceedings of the 25th ACM SIGKDD International Conference on Knowledge Discovery & Data Mining*. 1957–1965.
[14] Ilya Loshchilov and Frank Hutter. 2017. Decoupled weight decay regularization. *arXiv preprint arXiv:1711.05101* (2017).
[15] Ilya Mironov. 2017. Rényi differential privacy. In *2017 IEEE 30th computer security foundations symposium (CSF)*. IEEE, 263–275.
[16] Alec Radford, Jeffrey Wu, Rewon Child, David Luan, Dario Amodei, Ilya Sutskever, et al. 2019. Language models are unsupervised multitask learners. *OpenAI blog* 1, 8 (2019), 9.
[17] Da Yu, Saurabh Naik, Arturs Backurs, Sivakanth Gopi, Huseyin A Inan, Gautam Kamath, Janardhan Kulkarni, Yin Tat Lee, Andre Manoel, Lukas Wutschitz, et al. 2021. Differentially private fine-tuning of language models. *arXiv preprint arXiv:2110.06500* (2021).
[18] Tianyi Zhang*, Varsha Kishore*, Felix Wu*, Kilian Q. Weinberger, and Yoav Artzi. 2020. BERTScore: Evaluating Text Generation with BERT. In *International Conference on Learning Representations*. https://openreview.net/forum?id=SkeHuCVFDr
[19] Yizhe Zhang, Siqi Sun, Michel Galley, Yen-Chun Chen, Chris Brockett, Xiang Gao, Jianfeng Gao, Jingjing Liu, and Bill Dolan. 2020. DIALOGPT : Large-Scale Generative Pre-training for Conversational Response Generation. In *Proceedings of the 58th Annual Meeting of the Association for Computational Linguistics: System Demonstrations*. Association for Computational Linguistics, Online, 270–278. https://doi.org/10.18653/v1/2020.acl-demos.30
[20] Ming Zhong, Da Yin, Tao Yu, Ahmad Zaidi, Mutethia Mutuma, Rahul Jha, Ahmed Hassan Awadallah, Asli Celikyilmaz, Yang Liu, Xipeng Qiu, and Dragomir Radev. 2021. QMSum: A New Benchmark for Query-based Multi-domain Meeting Summarization. In *Proceedings of the 2021 Conference of the North American Chapter of the Association for Computational Linguistics: Human Language Technologies*. Association for Computational Linguistics, Online, 5905–5921. https://doi.org/10.18653/v1/2021.naacl-main.472